# Towards predicting Pedestrian Evacuation Time and Density from Floorplans using a Vision Transformer

Patrick Berggold, Stavros Nousias, Rohit K. Dubey, André Borrmann

Technical University of Munich, Germany

patrick.berggold@tum.de

**Abstract.** Conventional pedestrian simulators are inevitable tools in the design process of a building, as they enable project engineers to prevent overcrowding situations and plan escape routes for evacuation. However, simulation runtime and the multiple cumbersome steps in generating simulation results are potential bottlenecks during the building design process. Data-driven approaches have demonstrated their capability to outperform conventional methods in speed while delivering similar or even better results across many disciplines. In this work, we present a deep learning-based approach based on a Vision Transformer to predict density heatmaps over time and total evacuation time from a given floorplan. Specifically, due to limited availability of public datasets, we implement a parametric data generation pipeline including a conventional simulator. This enables us to build a large synthetic dataset that we use to train our architecture. Furthermore, we seamlessly integrate our model into a BIM-authoring tool to generate simulation results instantly and automatically.

## 1 Introduction

During the design and development process of a building, various experts from different disciplines are involved. Within the Architecture, Engineering, and Construction industry particularly, there is a multitude of disciplines – each making decisions individually – and thus impacting the decisions of other disciplines and the entire workflow in subsequent stages of the project. Consequently, errors or miscommunication in the early stages of the project, when the most critical decisions on the building performance are made, may lead to substantial temporal or economic expenses in later phases (Gervásio, 2014).

In recent years, Building Information Modelling (BIM) has gained massive popularity in facilitating collaboration among different disciplines starting from the early design phases (Borrmann, et al., 2018). BIM-authoring tools allow all project participants to collaborate simultaneously on a building's digital representation, and retrieve building information such as floorplans directly from the model itself. Therefore, as the project progresses, building models are gradually developed and refined, from immature conceptual layouts in the early design phases to increasingly complex entities. During this process, many design options need to be explored and evaluated by architects and engineers. The early integration of simulations into the workflow is thereby greatly beneficial, as simulations illustrate how decisions and design choices from one or multiple disciplines affect the physical reality of the project, which supports decision-making and relieves decision uncertainties (Abualdenien & Borrmann, 2019).

Amongst the many factors that influence the design phases of a building, the analysis of pedestrian flow dynamics plays a crucial role, in particular for large public buildings and transportation hubs. It enables project engineers to learn about essential information, such as planning escape routes for evacuation or preventing overcrowding situations, and therefore identify bottlenecks in the design of the building. Consequently, pedestrian simulators are essential tools that influence important building design choices. However, time-consuming tasks, such as running long-lasting, resource-intense simulations, are potential bottlenecks that may block the workflow and decision-making in the building design processes for various project partners. Furthermore, in practice, several cumbersome, manual steps are necessary to obtain simulation results from the BIM model, involving its export, e.g. into an IFC (Industry Foundation Classes) file, from which the semantics and geometry are extracted and converted into special simulation project files, to subsequently generate the results (Clever, et al., 2021).

To this end, we propose a data-driven approach that can realistically and instantly predict pedestrian densities over time, and total crowd evacuation time based on a building's floorplan, as well as simulation configuration, in substantially shorter time than conventional simulation. Furthermore, we integrate our model into a BIM-authoring tool to generate instant, fully automated results from the BIM model itself, circumventing any time-consuming, manual export or conversion steps, and opening possibility for an interactive exploration of the solution space. To showcase and evaluate our approach, we create a large and realistic synthetic dataset of evacuation scenarios, utilizing a conventional pedestrian simulator. As strong advocates of open-source development within the scientific community, we share our code (Dynamo and training scripts) on GitHub: `github.com/patrickberggold/PedSimAutomation`.

The paper is organized as follows: the next section discusses related work in pedestrian dynamics simulations and machine learning-based surrogate models. Subsequently, Section 3 describes the methodology of our approach. Section 4 covers the experimental evaluation, and details about the training and the resulting metrics. Finally, Section 5 concludes this article.

## 2 Related Work

### 2.1 Microscopic simulations

The simulation of pedestrian and crowd dynamics has been frequently discussed in the literature, and many models have been developed to accurately describe and predict locomotion patterns. In particular, microscopic, agent-based models are capable of accurately describing individual interactions and choices, since they describe pedestrians as agents, which are heterogeneous individuals that behave according to some pre-defined rules (Cheng, et al., 2014). Furthermore, microscopic models can be categorized into differential and non-differential models (Cristiani, et al., 2014), with the Social Force Model (SFM) (Helbing & Molnar, 1995) being the most popular representative of the former category. The SFM is based on a force equation that contains both repulsive and attractive forces, which incorporate the interactions between agents and obstacles, and enforce destination-driven behaviour. With respect to non-differentiable microscopic models, the Optimal Steps Model (OSM) (Seitz & Köster, 2012) combines the continuity of space with the natural stepwise movement of humans, in which the agents determine their path according to a utility function that incorporates different attractive and repulsive potentials, similar to the SFM.

### 2.2 Surrogate Modelling

While the aforementioned and several other conventional models generate sufficiently accurate simulation results, the generation of those results is typically time-consuming and potentially resource-intensive, depending on various factors such as the number of agents, simulation duration, number of interaction terms, etc. This motivates the application of surrogate models, which are capable of generating results of similar quality, but orders of magnitude faster. Given the impressive advances of machine learning (ML) techniques in recent years across many different disciplines, surrogate modelling based on ML seems to be the most promising candidate for instantly generating simulation results. For instance, in (Lehmberg, et al., 2020), the authors apply a surrogate model based on the Koopman operator to generate a crowd density time series in a simple bus station scenario. While their approach demonstrates the general advantages of employing a surrogate model, it is only applied to one simple use case, and thus, it is unclear how well this technique generalizes to more complex real-world scenarios. Better generalization is achieved in (Sohn, et al., 2020), where the authors employ a deep learning (DL) framework for predicting the aggregation of crowd densities over the entire simulation time. In particular, aggregated change is predicted via a Convolutional Neural Network (CNN) as a one-dimensional numerical number. While this approach successfully describes general crowd motion behaviour, it neglects simulation time prediction, and does not sufficiently quantify how and when congestions occur. In (Clever, et al., 2021), another CNN is employed for predicting the mean densities in realistic train station scenarios, which makes it possible to identify potential bottlenecks in the building design. However, the mean density does not provide any information as to when overcrowding situations occur, and how they develop. Finally, in (Clever, et al., 2022), a CNN is employed to solely predict evacuation time, also for train station scenarios, without including overcrowding analysis. Hence, in this article, we aim at combining previous premises, by instantly and accurately predicting pedestrian simulation results. Simultaneously, we remove any manual export and conversion steps involved in the result generation by integrating our approach into the overall BIM model for full automation.

## 3   Methodology

Our approach focuses on developing a DL framework that can accurately and instantly assess the building design quality inside the BIM model with respect to pedestrian dynamics. Therefore, we outline our methodology in the following. Due to the limited availability of public datasets, our first step involves generating a large synthetic dataset of evacuation scenarios, utilizing three parametric models to train our DL model. This dataset encompasses various combinations of geometry and simulation input parameters, facilitating robust generalization capabilities, as detailed below. Furthermore, we formulate density prediction as a classification task, as depicted in Table 1 and visualized in Figure 3. Lastly, we provide an in-depth description of our DL architecture.

### 3.1 Dataset generation

Our dataset is based on three parametric models, which resemble office buildings with different hallway layouts and room arrangements. As BIM-authoring tool, we use Autodesk's Revit (Autodesk, 2023), which is in practice a frequently used software for model-based building design. In Revit, the integrated parametric modelling system is called Dynamo, which has access to the Revit API; it is able to execute operations within Revit, being able to construct regular models, and access element properties and geometry. Moreover, it is possible to generate different variants of building types, solely by varying the input parameters of the parametric model. Each of the three parametric models (displayed in Figure 1) is generated from individual Dynamo scripts, serving as simple use cases for office building types in evacuation scenarios. The input parameters to those parametric models are of a purely geometrical nature. Namely, we create twelve different versions of each parametric model, based on different combinations of geometrical input parameters, namely **a)** floorplan length, **b)** floorplan width, **c)** corridor width, **d)** number of rooms, **e)** bottleneck inclusion, **f)** obstacle inclusion. While the first four parameters are numerical inputs, the last two are Booleans, meaning that bottlenecks and obstacles may or may not be included in the parametric model. If set to 'true', a bottleneck is placed in front of the escape area, such as a staircase, or obstacles are placed on the way to the escape area. Figure 1 visualizes three different parametric models and Figure 2 presents a flowchart of the dataset generation process in three steps. Step 1 of Figure 2 depicts both the absence and the presence of the bottleneck- and obstacle-option.

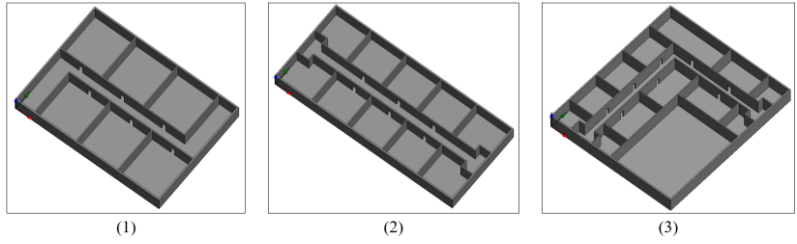

Figure 1: The three different parametric models, generated from individual Dynamo scripts.

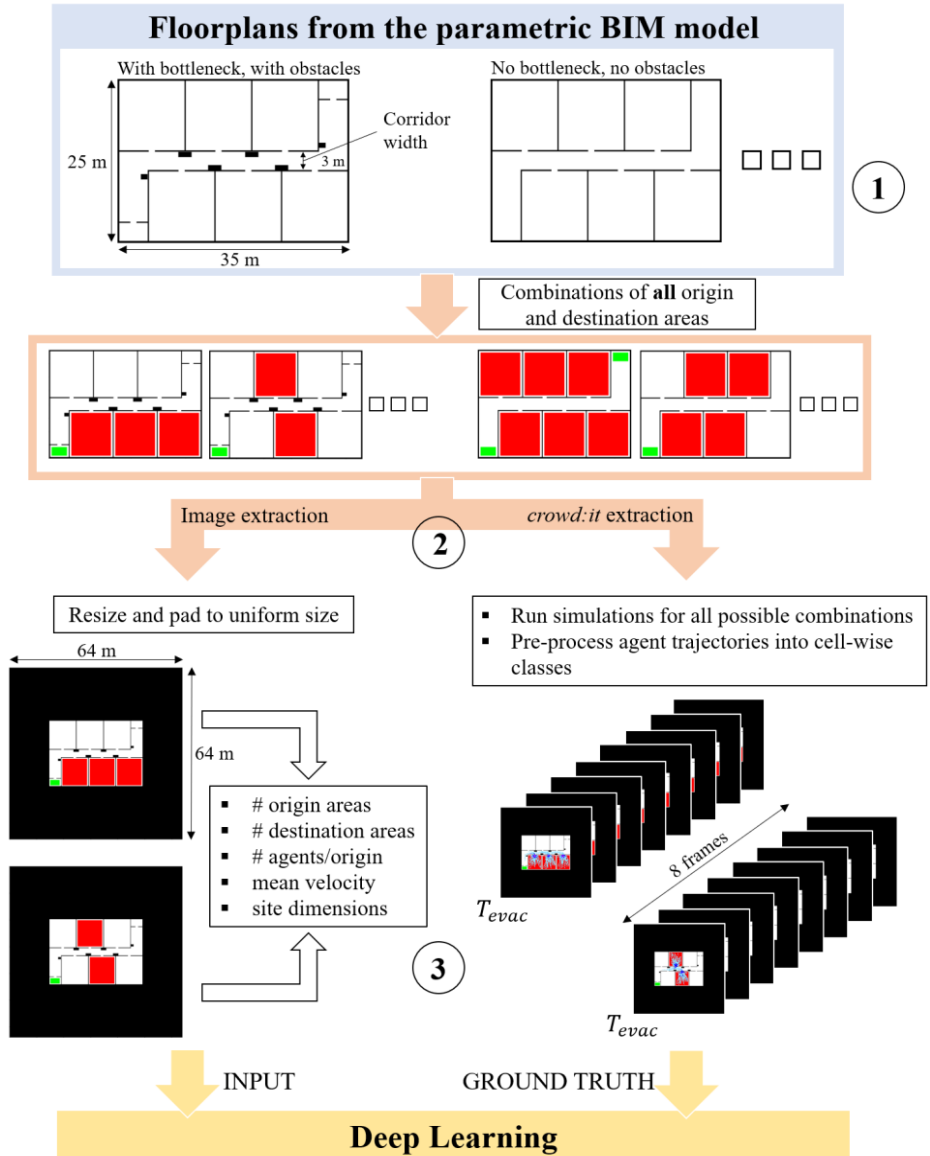

Figure 2: Flow chart of the dataset generation in three steps, including the floorplan export from Revit (1), the conversion into image and simulator format (2), and pre-processing for the deep learning pipeline (3).

As a conventional pedestrian simulator, we use the *crowd:it* simulator (accu:rate, 2023) to compute the coordinates of each agent over all timesteps via the OSM. In order to run pedestrian simulations in evacuation scenarios for the generated BIM models, we export the floorplans from Revit (Figure 2, step 1), and convert them subsequently to the XML-based *crowdit* format (Figure 2, step 2) for the simulator. With respect to pedestrian simulations, there is not just a variety of building geometries to consider, but also different simulation configurations that need to be covered. We apply numerous variations per building geometry through a second set of simulation input parameters, namely **a)** origin areas, **b)** destination areas, **c)** the number of agents per origin, and **d)** mean desired agent velocity. Specifically, the simulator requires the location and number of origin and destination areas. At the start of the simulation, a pre-defined number of agents is initialized into each of the origin areas, to subsequently move toward the nearest destination area as the simulation proceeds. After the simulation terminates (when the last agent has reached its goal), a four-dimensional table is created, consisting of timestamp, agent id, and agent coordinates $x$ and $y$. Thus, this table contains all agent trajectories, from initialization to termination.

In free space, each agent moves with its individual desired velocity, which is drawn from a Gaussian distribution that has the last simulation parameter as its mean, and a pre-defined default of 0.26 m/s as standard deviation. In accordance with the use case of evacuation scenarios, we place the origin areas into the rooms, and the destination areas at the ends of the hallways. For each of the 36 different building geometries, we run numerous pedestrian simulations. Specifically, each simulation represents a different combination of the simulation parameters. While the ranges for the first and seconds parameters depend on the room arrangements in the geometry, the second-to-last and last parameters range between 10, 20, and 30 agents per origin, and 1 m/s, 1.34 m/s (which is empirically the average, free-flow pedestrian velocity (Weidmann, 1992)), and 2.0 m/s mean desired agent velocity, respectively. One of the pedestrian simulations is shown in the upper half of Figure 3, with three origin areas (in red), one destination area (in green), 30 agents per origin, and 1.0 m/s mean velocity.

The combinations of all parameter values amount to a total of 130,356 simulations. On our CPU, an 11th Gen Intel(R) Core(TM) i7, the mean runtime of one simulation takes about 10.2 seconds, with a standard deviation of ca. 3.5 seconds. Note that for the given hardware, the runtime durations are likely to substantially scale up as more floorplan complexity is added and real-life use cases are included, e.g. multi-story buildings, more complex geometry (such as complicated convolutions), or larger floorplan dimensions. Additionally, the floorplans need to be exported and manually fed into the simulator in separate, consecutive steps, introducing another cumbersome and time-consuming chore in the simulation process, making it less efficient and prone to human errors.

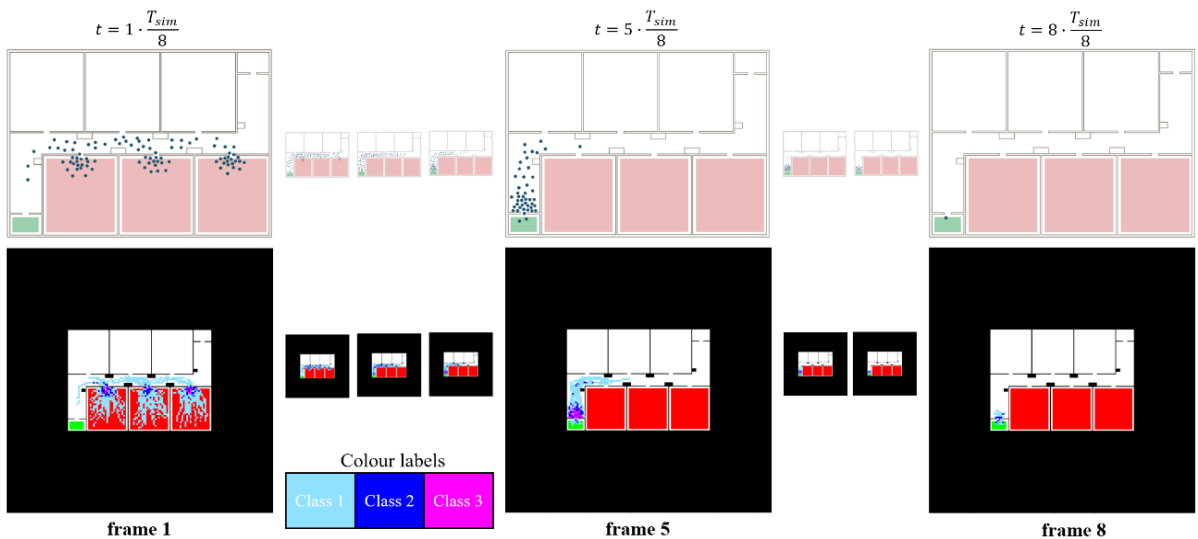

Figure 3: Conversion of the simulation results into eight consecutive frames. For each frame, the cell-wise densities are obtained and classified by counting the number of agents in each cell within the corresponding frame.

## 3.2 Density classification and pre-processing

We aim to build and train a deep neural network (DNN) that is capable of assisting the qualitative floorplan analysis based on the results of the pedestrian simulator. Accordingly, the dataset, which is used to train and evaluate our model, consists of numerous floorplan-to-simulation result mappings. Because of the major success of DNNs in image processing tasks, we supply the floorplans as images to the network, including the corresponding origin and destination areas. As displayed in Figure 2, steps 1 and 2, we export the floorplans from Revit and convert them to RGB image and simulator (.*crowdit*) format simultaneously.

For practitioners, the dynamics of macroscopic quantities that emerge from microscopic interaction rules, such as critical densities or pedestrian streams, are of particular interest, as they enable an accurate, qualitative analysis of when and where congestions occur, and how they develop over time. Therefore, our framework predicts the temporal development of agent densities plus total evacuation time (TET), based on floorplan and simulation parameter input. The variable shapes of the input and the transformation to macroscopic values in the output require a couple of pre-processing steps. Firstly, resizing and padding operations are applied to bring the floorplan images to a uniform size of 640x640 pixels. This image size translates into pre-defined maximum real-world dimensions of 64x64 meters. Thus, the floorplan images are resized according to their individual resolutions, centred into the square image, and padded with zeros (black colour). Afterwards, we subdivide the input image into a grid of sufficiently small cells. We choose a grid cell size of 4x4 pixels corresponding to 0.4x0.4 meters square, which is a common cell size for pedestrian experiments and simulations (Kouskoulis & Antoniou, 2017). To capture the dynamic development of the emerging macroscopic patterns, we subdivide each simulation into eight consecutive, equitemporal time frames, each of which with a duration of $\Delta t = \frac{T_{evac}}{8}$, where $T_{evac}$ is the TET, which is equal to the timestamp of the last agent reaching its destination (thus terminating the simulation). Within each frame, cell-wise density is determined by counting the number of agents $N_c$ per cell, as visualized in Figure 3. Since total simulation runtimes vary depending on geometry and parameter configuration, the respective frame durations vary as well, which is why the densities must be normalized to the frame interval $\Delta t$. Hence, we determine counts per time in each cell: $\tilde{\rho}_c = \frac{N_c}{\Delta t}$. In the simulations, the position of each agent corresponds to the centre of a circle representing the pedestrian's torso, whose diameter lies (by default) uniformly between 0.42 to 0.46 meters. We define the prediction of densities as cell-wise classification problem in the following way:

Table 1: Density class rules and descriptions

| Class | Rule | Description |
| --- | --- | --- |
| Class 0 | $\tilde{\rho}_c = 0$ | "No agents present" |
| Class 1 | $\tilde{\rho}_c \in (0, 0.4]\ s^{-1}$ | "Low density" |
| Class 2 | $\tilde{\rho}_c \in (0.4, 0.8]\ s^{-1}$ | "Increased density, congestion danger" |
| Class 3 | $\tilde{\rho}_c > 0.8\ s^{-1}$ | "High density, congestion" |

In summary, one sample in the dataset comprises the mapping from floorplan image and simulator configuration as input to the eight consecutive density heatmaps plus evacuation time $T_{evac}$ as output (Figure 2, step 3).

## 3.3 Deep learning architecture

The DNN predicts a series of consecutive density heatmaps plus TET from the floorplan input image and additional simulation information. Therefore, the model's architecture (see Figure 4) is inspired by the powerful image-to-image models that are capable of demonstrating state-of-the-art performance on semantic segmentation and image classification tasks. Namely, the basis of the model is the self- and cross-attention mechanisms that are commonly used in Vision Transformer (Dosovitskiy, et al., 2020) models. Firstly, the input images containing the floorplans are patched into 16x16 pixels and transformed into feature vectors via patch embeddings. Simultaneously, we supply additional simulation information in the shape of vectors to the network, consisting of the numbers of (1) origins, (2) destinations and (3) agents per origin, (4) the mean desired velocity, and (5) the site dimensions, meaning length and width of the floorplan in meters. This information is initially encoded using embeddings and linear layers, and subsequently concatenated and passed to a shared linear layer. In order to share and combine information between the image and information feature vectors, the encoder employs six consecutive, identical layers, each with a self-attention layer followed by a cross-attention layer. The first self-attention takes as input the 768-dimensional hidden states of the image encodings, and outputs hidden states that are the queries for the following cross-attention. Meanwhile, the cross-attentional modules also take as input the shared layer feature vectors, which are used as keys and values, and generates the input to the self-attention module in the subsequent encoder layer. The output of the encoder is used for density prediction, by decoding the hidden states with a series of transposed convolutional and max-pooling layers, followed by the UPerNet (Xiao, et al., 2018). We slightly modify its final layer by replacing it with a series of eight convolutional-batch norm layers that compute the cell-wise classification scores in each of the eight frames. Furthermore, the encoder states are also used as keys and values to another 3-layer cross-attentional modules, which takes as query input the output of the shared information layer. TET is predicted through a final linear layer with one-dimensional output.

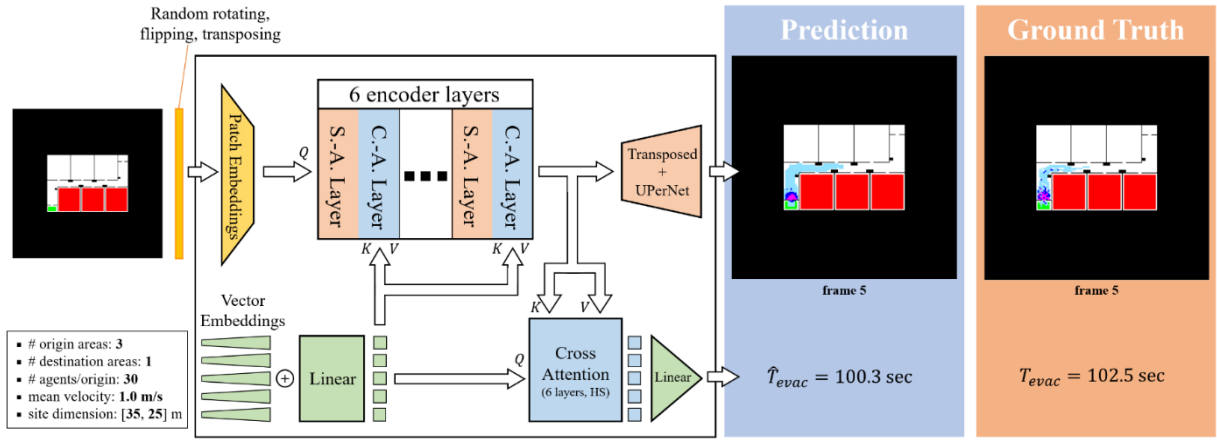

Figure 4: Overview of the DNN architecture, which receives as input the floorplan image plus additional simulation information and generates as prediction total evacuation time and eight density heatmaps (only frame 5 is shown here for clarity).

## 4 Experimental evaluation

### 4.1 Training

During training, the input images and density heatmap targets are flipped horizontally and vertically, transposed, and rotated around 90 degrees, all with probability of 0.5. We train the architecture from scratch, initializing all module weights according to Glorot (Glorot & Bengio, 2010). To cover both density and evacuation time prediction errors, we use a combined loss function, consisting of a classification and a regression error:

$$L_{total} = L_{evac} + \lambda_T \cdot L_{Tversky} = MSE(T_{evac}, \hat{T}_{evac}) + \lambda_T \cdot (1 - TI(y_{evac}, \hat{y}_{evac}))$$

The total loss $L_{total}$ is the sum of the regressive mean-squared-error (MSE) loss $L_{evac}$ with respect to TET $T_{evac}$, and the Tversky loss $L_{Tversky}$ (Salehi, et al., 2017) for cell-wise density classification over all time frames. An additional scaling coefficient $\lambda_T$ is introduced as hyperparameter to balance the two losses to facilitate simultaneous training.

The Tversky loss is chosen as classification loss to tackle the heavy class imbalance that exists in the density heatmaps, with the vast majority of cells being labelled as class 0 ( >95% in each frame). It originates from the Tversky index $TI$ (Tversky, 1977) that is an asymmetric similarity measure on two feature sets, for instance the predicted labels $P$ and ground truth labels $G$:

$$TI = \frac{|P \cap G|}{|P \cap G| + \alpha \cdot |P \backslash G| + \beta \cdot |G \backslash P|} = \frac{TP}{TP + \alpha \cdot FP + \beta \cdot FN} \text{ with } \alpha, \beta \geq 0$$

Given the true positives $TP$, the Tversky index measures the similarity between $P$ and $G$ by penalizing the false positives $FP$ against the false negatives $FN$ via the two hyperparameters $\alpha$ and $\beta$. Thus, the Tversky index lies between 0 and 1, with 1 being perfect classification. After several optimization runs, we settled on $\alpha = 0.1$ and $\beta = 0.9$. This means false positives are penalized substantially less than false negatives, prioritizing the minority classes.

We noticed that the model struggles when optimizing both classification and regression simultaneously in the early stages of the training, leading regularly to stagnation after only few epochs. Consequently, we pre-trained the model solely on the more complex task of cell-wise classification, neglecting the cross-attentional and linear modules associated with evacuation time prediction. The corresponding training and validation curves can be seen in Figure 5a. Subsequently, we trained the model on the simultaneous optimization task (see Figure 5b), while freezing all image processing weights in the first epoch (and unfreezing them afterwards).

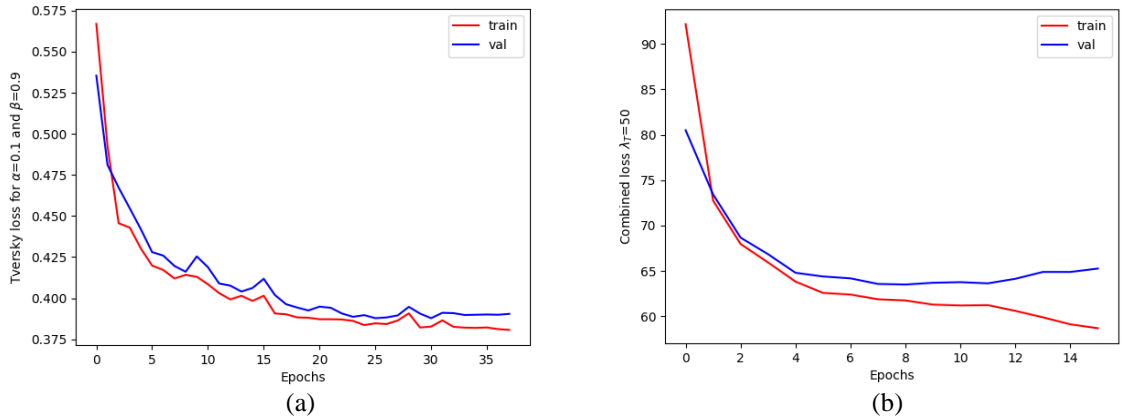

(a) (b)
Figure 5 (a) Pre-training curves of cell-wise density classification, (b) training curves with the combined loss $L_{total}$, consisting of the MSE and Tversky loss. Both trainings were conducted with the Adam optimizer, using an initial learning rate of 5e-4, the 'ReduceLROnPlateau' scheduler (patience=3) and early stopping.

## 4.2 Results

In order to interpret the training results, we evaluate the evacuation time and density heatmap predictions separately on unseen samples in the test set. For the classification task, a multi-class confusion matrix over the test set is displayed in Figure 6, summarizing the performance of the classification capability with ground truth labels along the y-axis, and the predicted labels along

x. Concerning evacuation time, we simply measure the mean absolute error $MAE$ between ground truth and prediction, and the corresponding relative error $RE$ related to the true evacuation time over all test set samples $N_{test}$. Altogether, the network predicts TET with more than 90% accuracy, which is a substantial increase to the results in (Clever, et al., 2022).

$$MAE = \frac{1}{N_{test}} \sum_i^{N_{test}} |T_{evac,i} - \hat{T}_{evac,i}| = \mathbf{4.65\ sec}$$

$$RE = \frac{1}{N_{test}} \sum_i^{N_{test}} \frac{MAE_i}{T_{evac,i}} = \mathbf{8.62\%}$$

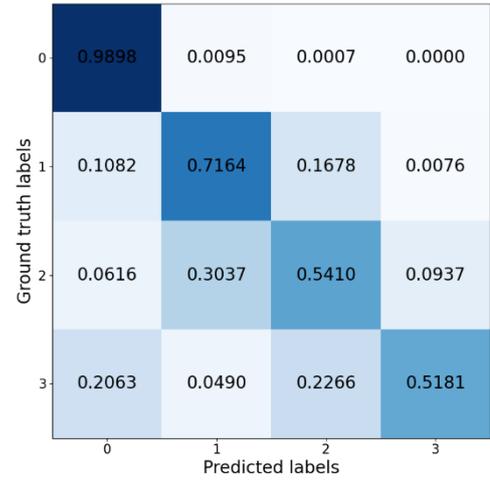

Figure 6: Multi-class confusion matrix of the density classification prediction.

### 4.3 Integration into Revit

Lastly, model inference is executed within Dynamo by integrating a customized Python environment into a Python block (named *DNN Forward pass* in Figure 7). Accordingly, the simulation parameters and site dimensions are supplied via the *Geometry* and *Simulation parameters* blocks, as well as the floorplan image that is generated from the rest of the Dynamo script. On our CPU, the network's forward pass generating the prediction takes roughly 1.7 seconds inside Dynamo, representing a significant speed-up compared to the conventional approach of undergoing several manual, laborious steps.

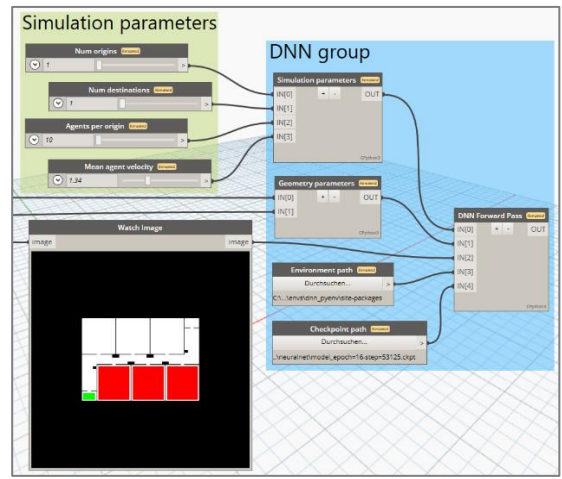

Figure 7: Integration of the trained model into Dynamo to fully automate the generation of simulation results from the BIM model.

## 5 Conclusion

In this article, we present an architecture capable of predicting both densities over time and total evacuation time realistically, based on floorplan information and simulator configuration. Thus, we show that our model is able to qualitatively assist the floorplan analysis, particularly in the early stages of the building design process. By integrating our model into the parametric modelling system Dynamo, we fully automate the generation of simulation results, achieving a significant speed-up compared to conventional approaches. An inherent constraint of our proposed methodology is the requirement of predetermined maximum floorplan dimensions for efficient model training with uniform image sizes. For larger floorplans, re-training the model is necessary, and for considerably smaller ones, the capabilities of cell-wise density prediction may deteriorate. Nevertheless, our methodology showcases a noteworthy improvement in efficiency by saving time and computational resources during the building design process.

## 6 Acknowledgments

This work was supported by mFUND – Bundesministerium für Digitales und Verkehr in Germany by funding the research project BEYOND. Furthermore, we acknowledge the support of Georg Nemetschek Institute for funding the project FORWARD as a continuation of the work presented in this paper.